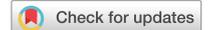

**OPEN**

# Winter wheat yield prediction using convolutional neural networks from environmental and phenological data

Amit Kumar Srivastava[1 ✉], Nima Safaei[2,6 ✉], Saeed Khaki[3,6 ✉], Gina Lopez[1], Wenzhi Zeng[4], Frank Ewert[1], Thomas Gaiser[1] & Jaber Rahimi[5]

Crop yield forecasting depends on many interactive factors, including crop genotype, weather, soil, and management practices. This study analyzes the performance of machine learning and deep learning methods for winter wheat yield prediction using an extensive dataset of weather, soil, and crop phenology variables in 271 counties across Germany from 1999 to 2019. We proposed a Convolutional Neural Network (CNN) model, which uses a 1-dimensional convolution operation to capture the time dependencies of environmental variables. We used eight supervised machine learning models as baselines and evaluated their predictive performance using RMSE, MAE, and correlation coefficient metrics to benchmark the yield prediction results. Our findings suggested that nonlinear models such as the proposed CNN, Deep Neural Network (DNN), and XGBoost were more effective in understanding the relationship between the crop yield and input data compared to the linear models. Our proposed CNN model outperformed all other baseline models used for winter wheat yield prediction (7 to 14% lower RMSE, 3 to 15% lower MAE, and 4 to 50% higher correlation coefficient than the best performing baseline across test data). We aggregated soil moisture and meteorological features at the weekly resolution to address the seasonality of the data. We also moved beyond prediction and interpreted the outputs of our proposed CNN model using SHAP and force plots which provided key insights in explaining the yield prediction results (importance of variables by time). We found DUL, wind speed at week ten, and radiation amount at week seven as the most critical features in winter wheat yield prediction.

Agricultural challenges aggravated by climate change and the increased food demand driven by the ever-growing population (expected to reach 9 billion by 2030)[1], highlight the importance of having a timely crop yield prediction model at a regional scale that can be used for better managing crops, ensuring food security, and improving policy and agricultural decision making. For this reason, a variety of approaches, ranging from process-based models to data-driven statistical algorithms, have been developed and applied to predict crop yield. However, since the process-based models for simulating the physiological mechanisms are constrained by the availability of data for parameterization, model calibration and validation[2], statistical methods offer promising alternatives and complementary tools. Machine learning (ML), is a practical statistical approach that has become popular due to big-data technologies and high-performance computing advancements. ML algorithms have created novel opportunities to aid farmers' decision-making[3,4] and inform actions in various real-world problems with or without minimal human intervention[5–8]. ML methods have the advantage of autonomously solving large nonlinear problems by making use of datasets compiled from different sources; they also provide a flexible and robust framework for making decisions and incorporating expert knowledge into the system based on data[9,10]. However, inconsistent spatial and temporal data about some of the production and management inputs (e.g., planting date, fertilizer application rate, crop variety- specific data) poses a great challenge to efficient training of the ML algorithms that needs to be addressed. Some algorithms are not intrinsically stable and might lead to inaccurate yield estimates while being applied on unseen datasets. Therefore, high prediction accuracy and computational speed, and good consistency among the results must be prioritized when using ML algorithms. Over the years, several ML algorithms have been used for crop detection and yield forecasting of certain crops in various locations, such as Convolutional Neural Network (CNN), Random Forests (RF), K-Nearest Neighbor (KNN), Least Absolute Shrinkage, and Selection Operator (LASSO) and Ridge Regression, Regression Tree (RT), Support Vector Machine (SVM), XGBoost, and Deep Neural Network (DNN)[11–17]. These methods have been systematically reviewed and discussed in the literature[9,18–20]. As an example, Van Klompenburg et al.[15] reported CNN, LSTM (long short-term memory networks), and DNN as the most often used deep learning (DL) algorithms for predicting crop yield. Among applications of ML algorithms for winter wheat yield prediction, Wang et al.[21] applied and compared six ML models (Ordinary Least Square (OLS), LASSO, SVM, RF, AdaBoost, and DNN) to predict the winter wheat yield within the growing season in the United States at county-level and ranked AdaBoost as the best algorithm ($R^2 = 0.86$, RMSE = 0.51 t ha$^{-1}$). Furthermore, Cao et al.[22] compared RF, as a traditional ML method, with three DL methods, i.e., DNN, 1D-CNN, and LSTM, to predict winter wheat yields in China, and concluded that all the tested methods performed well for this purpose ($R^2$ ranging from 0.83 to 0.90 and RMSE ranging from 0.56 to 0.96 t ha$^{-1}$). Since the performance of ML methods differ regionally, a comprehensive assessment of the accuracy of different methods for predicting the winter wheat yield values needs to be carried out separately for each major growing area. Here we concentrate on Europe, accounting for 32.7% of the total global wheat production, and particularly Germany, being the second-largest wheat producer in the European Union (EU) with 14.8% of the EU total wheat production[23]. Thus, this study aims to (1) benchmark the winter wheat yield prediction in Germany using state-of-the-art supervised ML algorithms; (2) propose a CNN-based architecture with a 1-D convolution operation to outperform other baselines in terms of accuracy; and (3) evaluate the effects of weather, soil and crop phenology variables on the prediction results and study the critical ranges of each feature on increasing/decreasing the yield output across different times of the year.

[1]Institute of Crop Science and Resource Conservation, University of Bonn, Bonn 53111, Germany. [2]Department of Business Analytics, Tippie College of Business, University of Iowa, Iowa, USA. [3]Industrial and Manufacturing Systems Engineering Department, Iowa State University, Ames, USA. [4]State Key Laboratory of Water Resources and Hydropower Engineering Science, Wuhan University, Wuhan 430072, China. [5]Karlsruhe Institute of Technology (KIT), Institute of Meteorology and Climate Research, Atmospheric Environmental Research (IMK-IFU), Karlsruhe, Germany. [6]These authors contributed equally: Nima Safaei and Saeed Khaki. ✉email: amit@uni-bonn.de; nima-safaei@uiowa.edu; skhaki@iastate.edu

## Data acquisition

The data analyzed in this study include yield value, weather, crop phenology, and soil features for the entire territory of Germany (271 counties) from 1999 to 2019. The data sources are provided in Supplementary Information section (Table S1).

**Weather data.** Temperature (minimum and maximum), radiation, precipitation, relative humidity, and wind speed data at a daily time scale were provided by the Deutscher Wetterdienst (DWD) for 21 years (ranging from 1999 to 2019) which were then interpolated to a 1 km grid scale using the method explained by Zhao et al.[24] and aggregated to weekly values at the level of NUTS3 (Nomenclature of Territorial Units for Statistics[25]) for input to the ML models. For aggregation to the level of NUTS3, area weighted average values were calculated using the agricultural land-use ratio in each grid cell. The calculations were based on CORINE Land Cover 2006 land-use data available at a resolution of 250 m[26]. The meteorological variables used in our analyses are wind speed, maximum and minimum temperature, relative humidity, precipitation, and solar radiation.

*Soil data.* Soil data was prepared and aggregated at DWD grids level using the major soil types (soil categories corresponding to the agricultural land-use categories as per CORINE Land Cover 2006). The source of the soil data is Germany's soil reconnaissance map available in 1:1,000,000 ratio (differentiated by BÜK1000N (BGR) land use)[27]. The soil variables included the volumetric (%) crop available water at permanent wilting point (LL), crop available water at field capacity (DUL), and saturation point (SAT), and bulk density (BD) available up to soil depth of 1.3 m. We used 1 km$^2$ as the spatial resolution of the soil data.

*Crop yield data.* Crop yield statistics of 271 counties at the subnational NUTS3[28] level for winter wheat was studied for years 1999 to 2019 across Germany. The regional database of Germany was used to extract NUTS3 crop yield data for the study[29].

*Crop phenology data.* Instances of winter wheat sowing, flowering, and harvest were accessed from the DWD phenology database[30]. DWD 1 km simulation grids were created by resampling the NUTS3 level data and joining grid centers to NUTS3. In the next step, we randomly sampled 50 weather and soil samples from each county and calculated their averages to extract representative meaningful values for weather and soil data (Fig. S1 in Supplementary Information).

We downsampled the daily weather data by taking the average and aggregating the feature values at the weekly level. The daily data is considered too detailed and granular, which would be problematic for knowledge discovery. Using weekly information reduced the dimension of the weather data with a 365:52 ratio, which significantly diminished the number of model parameters. Such preprocessing and downsampling the daily weather data to a weekly level is a common practice in yield prediction studies[11,12,31,32]. The combination of weekly meteorological, crop phenological, and soil variables across 271 counties in Germany from 1999 to 2019 resulted in a total of 5692 instances and 277 column features. The summary statistics of winter wheat yield (dependent variable) and all independent variables used in the study are described in Supplementary Information (Tables S2 and S3). Figure 1 shows a heatmap of correlation coefficients across all variables used in the study. For the weather data, the average output across the weeks is used in the graph. Based on Fig. 1, the yield value has the highest correlation with flowering and harvest day of the year and relative humidity. There is a high correlation observed between relative humidity and sowing, flowering, and harvest day, and radiation values.

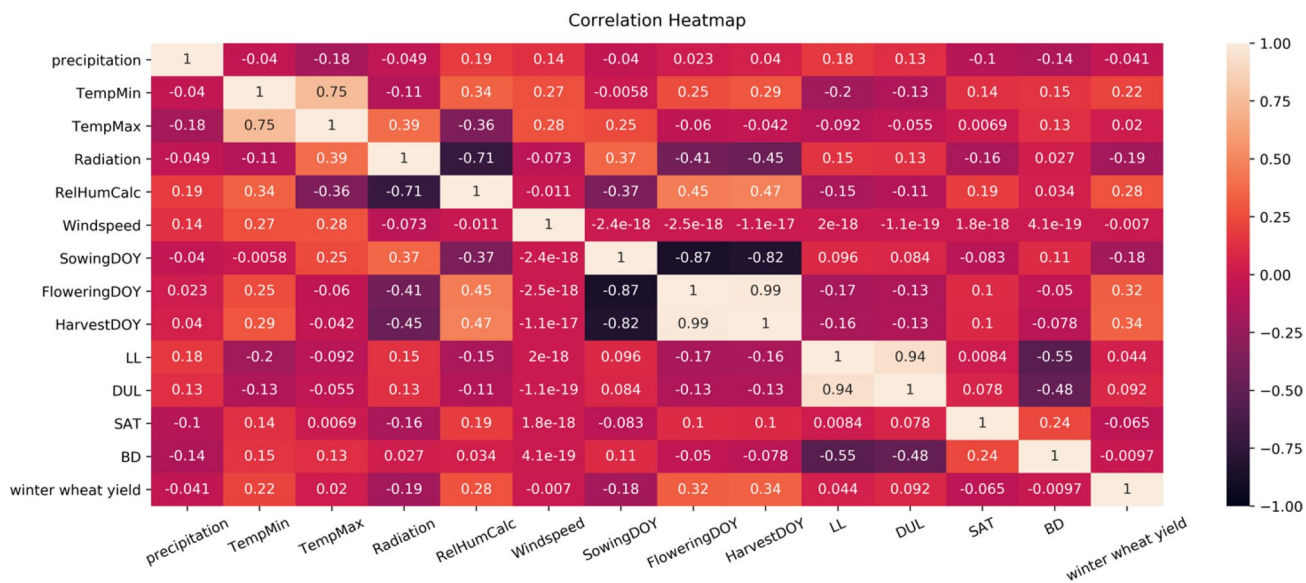

**Figure 1.** Pearson correlation heatmap of weather, phenology, soil, and yield data; darker colors show high correlation (close to 1), and lighter colors show low correlation (close to 0).

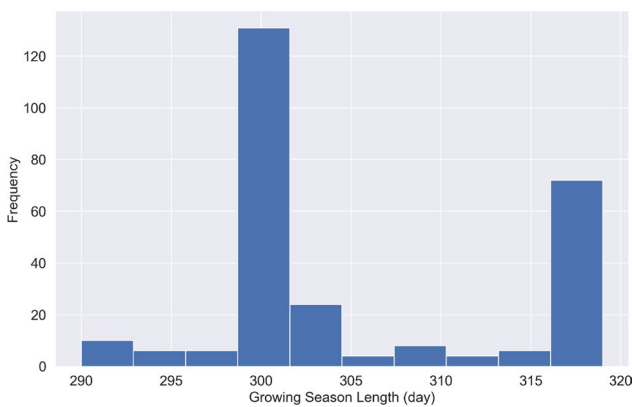

**Figure 2.** Frequency distribution of growing season length (from sowing to harvest) across the 271 counties in 2019.

There are different sowing/harvest dates for various counties, which results in different growing season lengths. The histogram of growing season length across all counties in 2019 is shown in Fig. 2.

Based on the summary statistics results of input variables in Table S3, they have varying scales and a diverse range of values. To avoid the intrinsic influence and dominance of a single feature over prediction results, all feature values are standardized using the z-score normalization technique [Eq. (1)]. Using this method, all features are rescaled to have the properties of a standard normal distribution. There are also other benefits associated with standardizing the feature values, such as enhancing the models' numerical stability and improving the training speed.

$$X_{scaled(i)}^{j} = \frac{X_{i}^{j} - \mu^{j}}{\sigma^{j}} \tag{1}$$

$X_{i}^{j}$ refers to the i$^{th}$ instance of the $j^{th}$ feature ($j = 1$ to $K$ where $K$ is the total number of independent variables). $\mu^{j}$ and $\sigma^{j}$ are the mean and standard deviation of the $j^{th}$ feature.

## Methods

**Description of ML models.** *Proposed method (CNN).* The proposed method of this study combines convolutional neural networks and fully-connected (FC) neural networks for winter wheat yield prediction. The convolutional neural network part of the model takes in the weather variables measured through the growing season as input and captures their temporal and nonlinear effects using 1-dimensional convolution operations. All six weather variables go into the same CNN model separately, and we concatenate their corresponding output of the CNN. The soil and phenology data go to a fully-connected neural network with two layers. We combine the high-level features of the CNN with the output of the fully-connected neural network for soil and phenology data. Finally, the combined features go to three other FC layers before the final yield prediction. The modeling architecture of the model is shown in Fig. 3.

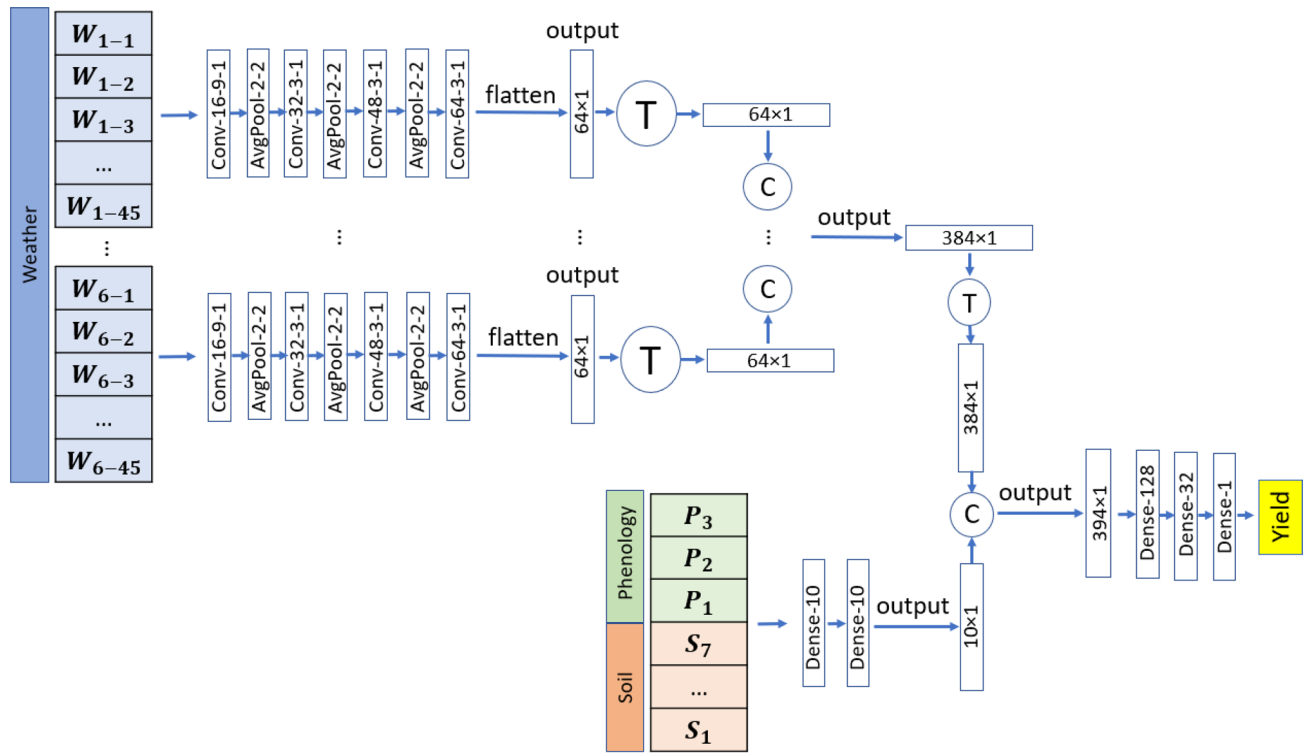

**Figure 3.** Network architecture of the proposed CNN. The convolutional layer parameters are shown as "(convolution type— number of filters—kernel size—stride size)". Conv, AvgPool, and Dense stand for standard convolutional, average pooling, and fully connected layers, respectively. The padding type is "valid" for all layers. ⊙ shows matrix concatenations and ⊤ refers to matrix transpose. The activation function of all layers except the last layers is ReLu.

*Random Forest (RF).*   RF makes use of multiple trees and uses "bagging" for ensembling decision trees into a single model[33]. It uses bootstrapping to randomly select K samples from the original dataset. These training samples are used to generate decision trees for developing the RF model.

*K-Nearest Neighbor (KNN).*   K-Nearest Neighbor (KNN) is a non-parametric method, introduced by Cover and Hart[34], used for classification and as well as regression type problems. It computes the distances of the testing instance from the training data to identify the closest neighbors and predict the output. The Euclidean distance equation (Eq. 2)[35] was used in this study to calculate the distances between the data points.

The Euclidean distance equation can be represented as Eq. (2) where,

$$d(p,q) = \sqrt{\sum_{i=1}^{n}(q_i - p_i)^2} \tag{2}$$

*d* is the Euclidean distance; *p* and *q* are the data points consisting of various dimensions; *n* is the total number of data points, and *i* is an index number.

*LASSO and Ridge Regression.*   Lasso and Ridge Regression are two distinct methods of reducing the variance of linear regression models by shrinking their coefficients. The Least Absolute Shrinkage and Selection Operator (LASSO) is a regularization method that can set the coefficients of some features to zero and consequently drop them from the model[36]. In this manner, Lasso seems like a feature selection algorithm. In LASSO, the L1 regularization is added to the objective function of the linear regression model, which shrinks coefficients toward zero. On the other hand, ridge regularization (L2 regularization) reduces coefficients toward zero, but it never sets any of them to zero[37].

*Regression Tree (RT).*   A tree is created by splitting the root node of the tree into various partitions that constitute the successor nodes. The process is repeated recursively on each partition known as binary recursive partitioning[38]. Initially, all data points are in the same group and the algorithm partitions the data into binary branches. Across the possible binary splits in each of the tree levels, the tree algorithm used in this paper chooses the split with the minimum sum of squared differences from the mean in the two partitions. The splitting continues till every node becomes a terminal node with a pre-defined minimum node size or if the node's sum of squared difference from the mean becomes zero (whichever is reached earlier in the tree).

*Support vector regression (SVR).* SVM uses a classification algorithm for predicting a continuous variable by fitting the best line within a threshold of values (epsilon-insensitive tube)[39]. This tube contains the margin of error the model can have. The distance between the data points and the tube is measured and labeled as the support vectors. The objective function of the optimization problem is shown in Eq. (3).

$$Min \frac{1}{2}||w||^2 + C \sum_{i=1}^{n} |\epsilon_i| \qquad (3)$$

where $w$ stands for the distance between the support vectors; $C$ is a trade-off parameter defined by the user and $\epsilon_i$ is the margin of error.

*XGBoost.* XGBoost stands for extreme gradient boosting (improved version of gradient boosted decision trees based on speed and performance), and its inception dates back to a project carried out by Tianqi Chen in 2016[40]. XGBoost is an ensemble of decision trees; it consists of a sequential build-off from various decision trees where each tree works to enhance the performance of the prior tree. In XGBoost, the training of each tree is parallelized, and this significantly boosts the training speed. XGBoost has been widely used for crop yield prediction[11,14,41,42].

*Deep Neural Networks (DNN).* DNNs are artificial neural networks with multiple levels of abstraction, which can learn the underlying representation of data without the need for handcrafted features[43]. The DNN model used in this paper is a feedforward neural network with multiple hidden layers. DNNs apply a nonlinear function to the output of each hidden layer which makes them highly nonlinear. DNN models are trained with gradient-based optimization methods to minimize the desired loss function for the task they are used. DNN models have shown great success by outperforming other traditional machine learning methods in predicting the crop yields[12,15,16,44–46].

The traditional neural networks with single hidden layer have also been widely used to estimate DUL and permanent wilting point[47,48]. This method was inspired by human brain structure[49] and can learn data pattern from the training dataset and save them as weighted connection of nodes. After learning, the network recognizes the pattern in the given dataset and predicts the output[50].

We used Python programming language for writing the scripts, Python visualization libraries such as Matplotlib and Seaborn for drawing the graphs, and the machine learning and deep learning libraries of Python, including scikit-learn and TensorFlow for training and testing the prediction models. We also used QGIS software for plotting the spatial graphs.

**Model interpretability.** Recently, there has been a massive inclination toward applying interpretation tools on black-box models such as tree-based ensembles (e.g., Random Forest and Gradient Boosting Trees) and deep learning models (e.g., DNNs, CNNs, and RNNs) for explaining the yield prediction results[11,11,14,51]. In general, deep neural network models lead to more accurate predictions than other interpretable machine learning models, such as Linear Regression. However, the black-box nature of deep learning models does not disclose the reasons (features) that are important in predicting crop yield values and whether these features are decreasing or increasing the yield outputs. Therefore, in this study, we apply SHAP as a post-hoc method to calculate Shapley values and explain the predictions of our proposed CNN model. In this way, the model achieves both optimal accuracy and decent interpretability. Shapley values were created based on the game theory[52]; they can be used to show the contribution of each feature in the model's outcome. In computing the Shapley values, the mean marginal effect of feature values on the prediction results is analyzed by evaluating feature permutations for all features to calculate their associated Shapley values. More important features are assigned with greater Shapley values due to their higher contribution to the model results. This paper uses an implementation of Kernel SHAP called KernelExplainer, an efficient, model-agnostic method for estimating SHAP values for any model.

**Input combinations and parameter calibration.** We used a combination of weekly meteorological, crop phenological (sowing, flowering, and harvest dates), and soil variables across 271 counties in Germany from 1999 to 2019 to train and test our ML models. This resulted in a total of 5692 instances and 277 column features. We tested various hyperparameter domain values on the validation set to achieve the best hyperparameter setting resulting in the lowest error on the test data. The validation dataset was chosen randomly using cross-validation from the data between 1999 and 2016. The best parameters setting was found using a grid or randomized search cross-validation with 1000 iterations on a 3-fold cross-validation setting (3000 trials for each model). The range of tested values for hyperparameters is chosen based on domain knowledge. The tested hyperparameters and the resulting best hyperparameter estimations for baseline models are shown in Supplementary Information (Table S4). The architecture and hyperparameters of the CNN model are described in Fig. 3.

**Model evaluation.** The performance of prediction models was evaluated using mean absolute error (MAE) [Eq. (4)], root means square error (RMSE) [Eq. (5)], and correlation coefficient (r) [Eq. (6)] metrics:

| | KNN | Random Forest | XGBoost | Lasso | Ridge | Regression Tree | SVR | DNN | Proposed CNN |
|---|---|---|---|---|---|---|---|---|---|
| Training 1999 to 2016 (RMSE) | 0.33 | 0.18 | 0.45 | 0.57 | 0.56 | 0.46 | 0.33 | 0.41 | 0.48 |
| Test on 2017 (RMSE) | 0.88 | 0.79 | 0.73 | 1.19 | 0.91 | 0.85 | 0.86 | 0.80 | 0.66 |
| Training 1999 to 2017 (RMSE) | 0.38 | 0.32 | 0.45 | 0.61 | 0.59 | 0.54 | 0.38 | 0.44 | 0.47 |
| Test on 2018 (RMSE) | 1.06 | 0.90 | 1.05 | 1.23 | 1.32 | 0.95 | 1.04 | 1.00 | 0.84 |
| Training 1999 to 2018 (RMSE) | 0.44 | 0.42 | 0.52 | 0.67 | 0.65 | 0.59 | 0.44 | 0.43 | 0.50 |
| Test on 2019 (RMSE) | 1.07 | 0.81 | 0.77 | 1.27 | 1.38 | 0.98 | 1.02 | 0.74 | 0.64 |
| Training 1999 to 2016 (MAE) | 0.25 | 0.14 | 0.35 | 0.44 | 0.44 | 0.35 | 0.24 | 0.32 | 0.38 |
| Test on 2017 (MAE) | 0.67 | 0.64 | 0.59 | 1.03 | 0.75 | 0.69 | 0.71 | 0.66 | 0.52 |
| Training 1999 to 2017 (MAE) | 0.27 | 0.23 | 0.38 | 0.47 | 0.46 | 0.41 | 0.27 | 0.35 | 0.37 |
| Test on 2018 (MAE) | 0.88 | 0.73 | 0.84 | 1.03 | 1.09 | 0.78 | 0.82 | 0.81 | 0.71 |
| Training 1999 to 2018 (MAE) | 0.30 | 0.29 | 0.41 | 0.51 | 0.50 | 0.44 | 0.29 | 0.34 | 0.39 |
| Test on 2019 (MAE) | 0.84 | 0.66 | 0.59 | 1.00 | 1.08 | 0.81 | 0.85 | 0.59 | 0.50 |
| Training 1999 to 2016 (r) | 0.94 | 0.99 | 0.89 | 0.81 | 0.82 | 0.89 | 0.94 | 0.91 | 0.90 |
| Test on 2017 (r) | 0.30 | 0.62 | 0.63 | 0.41 | 0.40 | 0.57 | 0.35 | 0.60 | 0.65 |
| Training 1999 to 2017 (r) | 0.92 | 0.95 | 0.88 | 0.78 | 0.80 | 0.83 | 0.92 | 0.90 | 0.88 |
| Test on 2018 (r) | 0.30 | 0.52 | 0.46 | 0.15 | 0.12 | 0.48 | 0.28 | 0.39 | 0.78 |
| Training 1999 to 2018 (r) | 0.89 | 0.91 | 0.84 | 0.73 | 0.75 | 0.80 | 0.90 | 0.90 | 0.87 |
| Test on 2019 (r) | 0.49 | 0.67 | 0.71 | 0.44 | 0.44 | 0.57 | 0.45 | 0.73 | 0.81 |

**Table 1.** Training and validation results (RMSE, MAE, and r (correlation coefficient)) of the baseline machine learning models and the proposed CNN model to predict winter wheat yield.

$$MAE = \frac{\sum_{i=1}^{n} |y_i - \hat{y}_i|}{n} \tag{4}$$

$$RMSE = \sqrt{\frac{\sum_{i=1}^{n} (y_i - \hat{y}_i)^2}{n}} \tag{5}$$

$$r = \sqrt{1 - \frac{\sum_{i=1}^{n} (y_i - \hat{y})^2}{\sum_{i=1}^{n} (y_i - \bar{y})^2}} \tag{6}$$

$\hat{y}_i$, $\bar{y}_i$ and $y_i$ stand for the predicted, average, and ground truth target values, respectively. n stands for the total number of data points.

## Results

**Prediction results.** Using the best hyperparameter settings achieved through hyperparameter tuning, three models are trained and validated for each of the machine learning methods in the study. The train-test splitting is done using a non-random hold-out method and is based on the time variable to ensure the valid training of the models for predicting future yield and avoiding retroactive prediction. For enhancing the reliability of the training and testing results, one model is trained on the data between 1999 to 2016 (training data) and tested on the data of 2017 (test data), the other model is trained using the data between 1999 to 2017 (training data) and tested on 2018 (test data), and the last model is trained on the data between 1999 to 2018 (training data) and tested on the data of 2019 (test data). The prediction performance results based on the train and test dataset are shown in Table 1.

Based on the results shown in Table 1, our proposed model (CNN) outperformed other tested methods to a varying extent due to its highly nonlinear structure as well as the ability to capture the temporal dependencies of weather data. XGBoost and DNN models showed comparable performance, while DNN was slightly better for some test years and XGBoost was marginally better during the others. DNN outperformed other baseline models (except XGBoost) based on all validation metrics as the DNN model is highly nonlinear and automatically discovers the relationship between the input data and the yield by extracting relevant features from the input data. The linear models, i.e., Lasso and Ridge Regression had a weak performance compared to nonlinear models due to their linear structure, which fails to capture the nonlinear effects of weather and soil conditions. XGBoost outperformed Random Forest, and Regression Tree models during most of the test years due to the use of gradient boosted trees which significantly improves its performance. KNN and SVR models had comparable performance, and they had a higher prediction accuracy than linear models due to capturing the nonlinear structure of variables. Although extensive hyperparameter tuning was conducted to find out the best set of hyperparameters leading to the minimum cross-validated prediction error, there is still a slight overfitting problem existent in the prediction results of some baseline models such as KNN, SVR, and Random Forest.

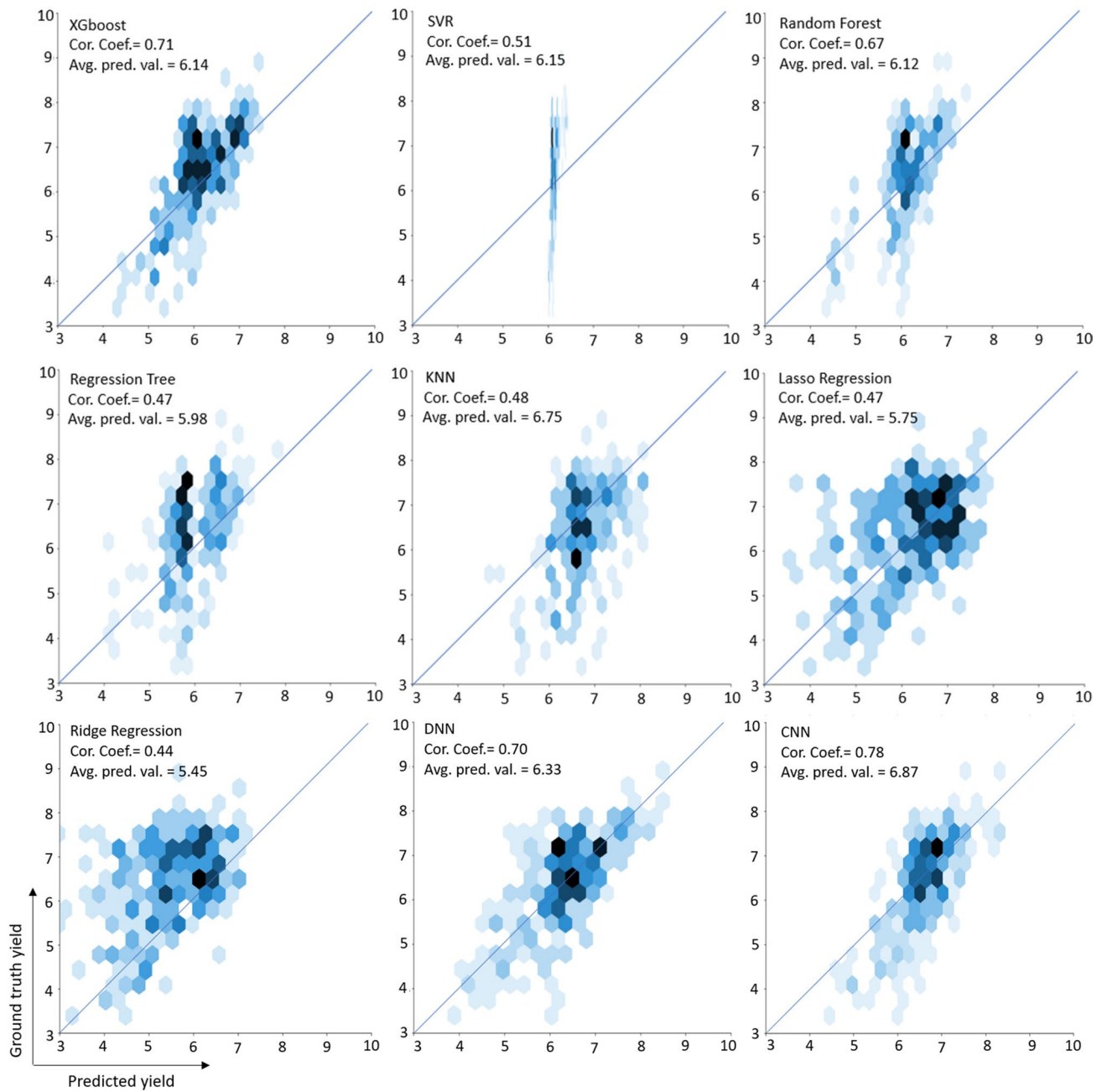

**Figure 4.** Hexagonal plots of the predicted winter wheat yield vs. ground truth yield values for the nine machine learning models in the test year of 2019.

We plotted the ground truth yield and predicted yield values of all the models using hexagonal plots in Fig. 4 to see how close the prediction results are to the ground truth values. Data values in the hex-binned plots are covered with hexagon arrays; the color of each hexagon is based on the number of observations it covers; darker hexagons show higher densities of data points, and lighter hexagons are representatives of lower densities. The deviation of the data points from the 1:1 line shows the distribution of residuals (for better observation of residual distribution, refer to Fig. 5). As shown in Fig. 4, for XGBoost, DNN, and CNN models, the predicted and ground truth values have a significantly stronger association compared to the other models, with CNN having the strongest association [XGBoost (r = 0.71), DNN (r = 0.70), and CNN (r = 0.78)]. In other words, more than 60 percent of the variability in the ground truth values can be explained by the predicted results of the CNN model ($R^2 = 0.61$). Based on Fig. 4, KNN, Lasso Regression, and Regression Tree have similar performance with respect to the correlation coefficient between the ground truth and the predicted yield value; seemingly, Ridge Regression has the worst performance as most of the predicted value hexagons are off-diagonal (It also has the lowest correlation coefficient among the models). The range of predicted yield for the SVR model is very narrow, which shows that the SVR model predicts most of the instances between 6 and 7 tons ha$^{-1}$ and cannot be a reliable predictor for yield values.

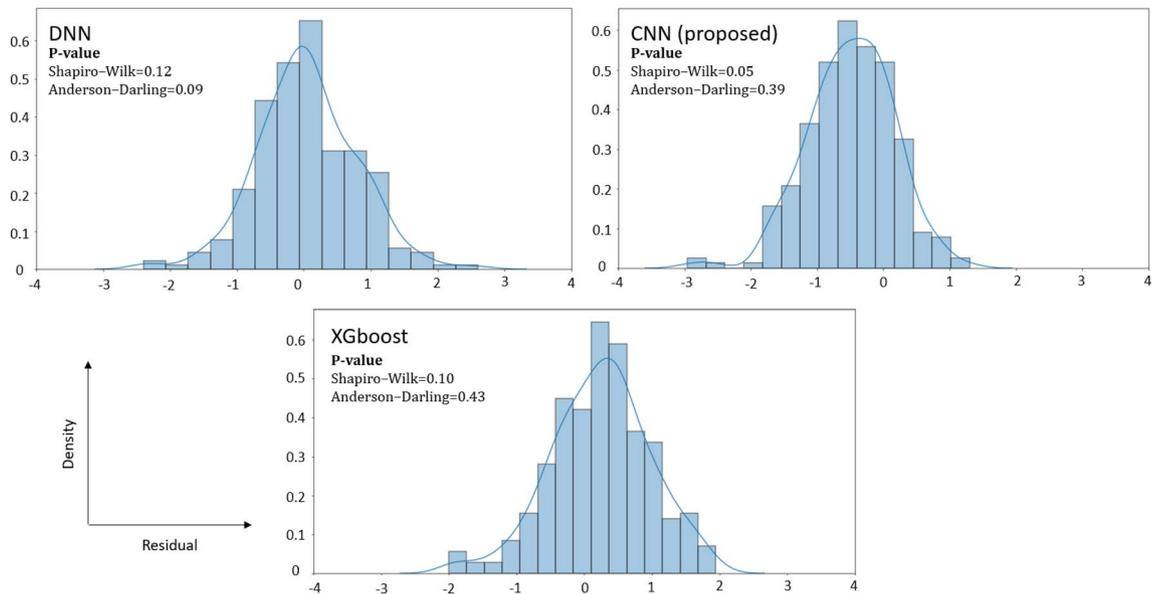

**Figure 5.** Distribution of prediction errors of XGBoost, DNN, and CNN models.

Although the primary goal of predictive modeling is to minimize the validation error and make the most accurate predictions, performing residual analysis would help us choose the best-performing model with more confidence. Figure 5 shows the distribution of prediction errors for the top three most accurate models observed in Table 1 (XGBoost, DNN, and CNN) using the test data (2019).

Based on Fig. 5 and the p-value results of Shapiro–Wilk and Anderson–Darling normality tests, we do not have sufficient information to reject the null hypothesis (normal distribution) for XGBoost, DNN, and CNN (p-value = 0.05; borderline); thus, we conclude that the prediction residuals are approximately normally distributed for these models. Again, in this paper, as we are focused on conducting predictive modeling using machine learning techniques, "we treat the data mechanism as unknown" as the data modeling assumptions of inferential statistics are less of a concern[33].

The spatial distribution of the prediction error for winter wheat yield using the proposed CNN, DNN, and XGBoost models across the 271 counties in 2019 is shown in Fig. 6a–c. Fig. 6d shows the distribution of observed yield values across different counties in the test year (2019). This figure helps identify counties with higher error percentage values, and diving deeper into this information could reveal helpful patterns to improve the data collection quality in these regions. Based on the CNN model results, the error percentage varies from 0.1 to 77.6% across the counties in 2019. Counties with higher prediction errors are primarily located in the eastern and north-eastern part of Germany; central counties represent lower error values. Although the range of the percentage error values is higher in the proposed CNN model than the DNN and XGBoost models, based on Table 1, the overall performance of the proposed CNN model is significantly better than the mentioned baselines. The prediction percentage error is calculated using Eq. (7). $A_i$ and $P_i$ stand for the actual and predicted yield values in county i.

$$Percentage\ error_i = |\frac{A_i - P_i}{A_i}| \times 100\% \tag{7}$$

**Analysis.** *Feature importance explanation.* We have predicted the yield using weekly weather data, soil conditions, and crop phenology variables. In this section, we explain the importance of each feature in the predictions made by our proposed CNN model to find the relative importance of features. Weather and soil data are generally represented by numerous variables, which do not possess equal effects or importance in yield prediction. Thus, it is of paramount importance to find more important features and help farmers and agronomists focus on those critical features that are more important in increasing the yield values than the others. In this paper, we used SHAP as a post-hoc method to explain the prediction outputs and find the input variables that have a higher contribution to the model outputs. Features with larger/smaller Shapley values show a higher/lower contribution to the predicted yield outputs. The average absolute Shapley value per feature across the data is calculated to find the global importance. The results are shown in Fig. 7. Red bars show features that positively correlate with yield impact (thus positively impacting the yield value) and blue bars show the features that negatively correlate with yield impact (thus negatively impacting the yield value).

In addition, to further dive into understanding the feature effects and their importance, we have created a series of force plots; the plots for some important features (refer to Fig. 7) in yield prediction are shown in Supplementary Information (Figs. S2–S5). They explain the contribution of several important features to the yield prediction values. The vertical axis in these plots shows the predicted yield value (in tons ha⁻¹), and the horizontal axis refers to the selected feature values. The intercept of the vertical and horizontal axis shows the average of the prediction outputs. Parts of the plots in red and blue indicate the range of feature values that positively and negatively affect the yield values, respectively. For instance, Fig. S2 shows that DUL lower than 0.27 cm³ cm⁻³ negatively contributes to the yield values. Also, Fig. S3 shows that higher radiation values around week seven after wheat planting negatively affects the yield values. Fig. S4 suggests that maximum temperature around week eight after wheat planting has a positive relationship with the yield values. Based on Fig. S5, higher precipitation amounts around week 49 negatively affects yield values.

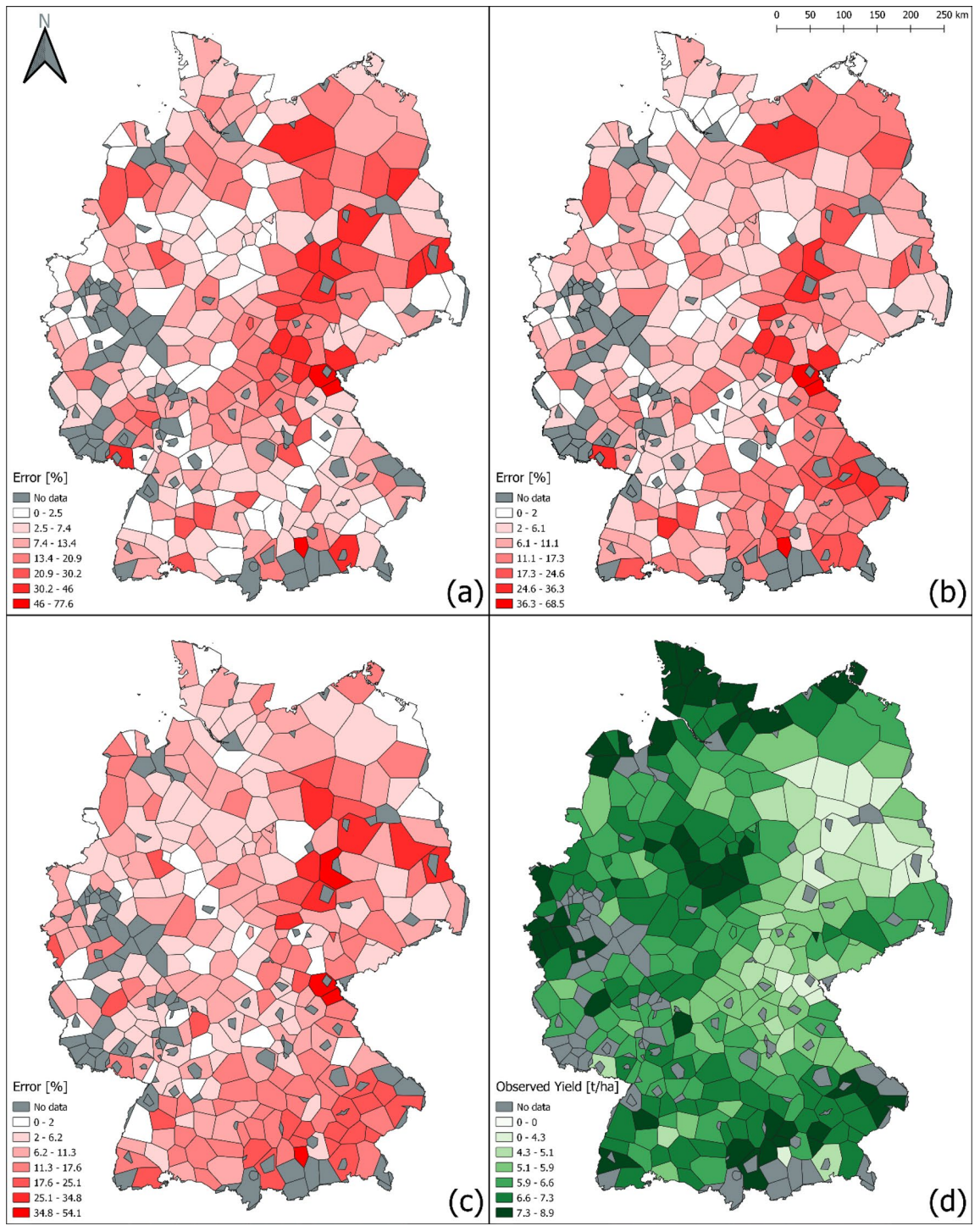

**Figure 6.** Percentage prediction error in winter wheat yield using (**a**) CNN, (**b**) DNN, (**c**) XGBoost models, and (**d**) distribution of observed yield (ground truth) in 271 counties. The counties in dark gray color are indicative of non-availability of ground truth data in that county in 2019. Figures were created using QGIS software version -3.18.3, https://qgis.org/en/site/.

Additionally, using SHAP explanation force plots, three instances with the highest, median, and lowest yield values were selected to explain the effects of their features on the predicted yield (Fig. 8). Force plots are effective in interpreting the prediction value of the model in critical instances. The contribution of each feature to the output predicted value is shown with arrows with their force associated with the Shapley values. Red arrows indicate features increasing the prediction results (i.e., yield values) to reach the predicted value (output value). Blue arrows show features decreasing the prediction values to reach the same output value. The arrows with positive and negative effects on yield values compensate on a point which is the prediction (output) value. The base values in Fig. 8 are the mean of the ground truth yield values over the entire test dataset.

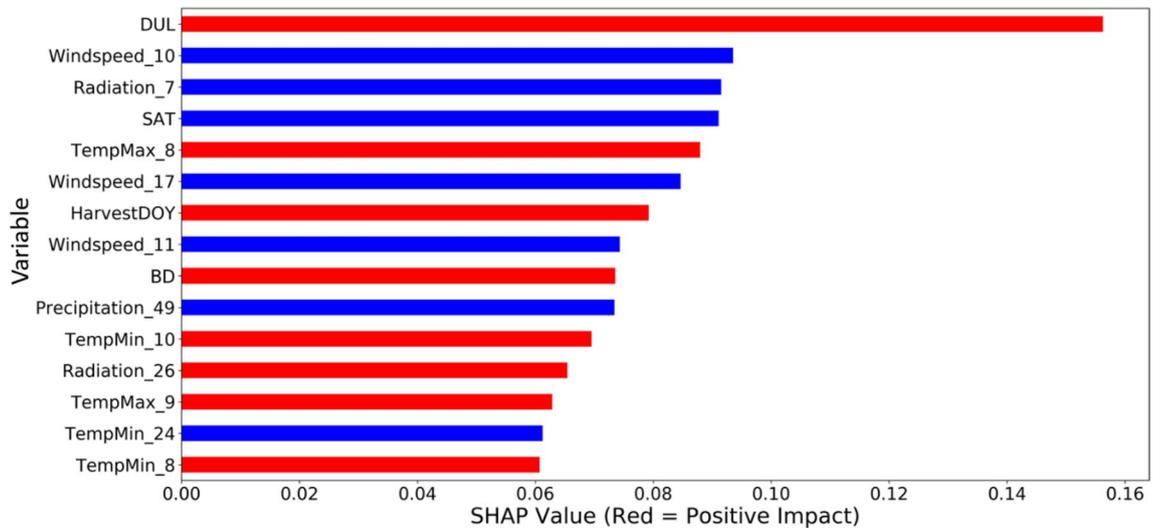

**Figure 7.** Shapley values for the top 15 features in winter wheat yield prediction in 2019 (For the weather variables on the vertical axis, the variable name refers to the title of the weather variable plus the week of the year the information was recorded, separated by an underscore sign) (*HarvestDOY* Harvest day of the year, *LL* crop available water at permanent wilting point, *SAT* crop available water at the saturation point, *DUL* crop available water at the field capacity, *BD* soil bulk density).

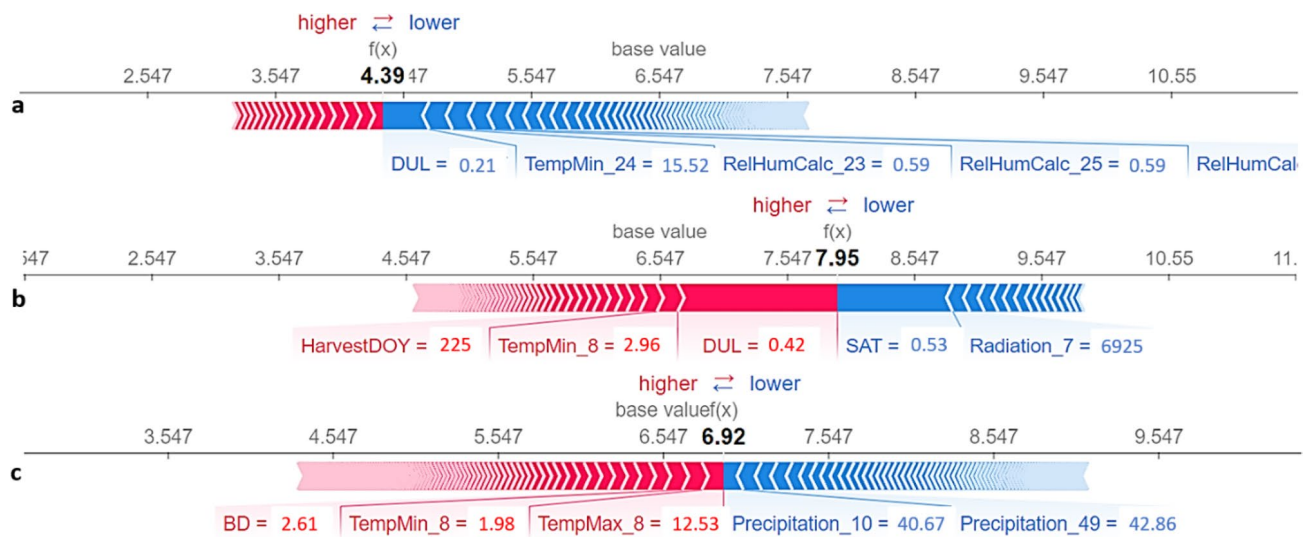

**Figure 8.** Force plots to explain the predicted yield values for instances with the minimum (**a**), maximum (**b**), and median (**c**) yield values in 2019.

Fig. S6 (in Supplementary Information) shows the estimated effects of weather components on the predicted winter wheat during 45 weeks of 2019. Out of the six weather variables examined, the wind speed, temperature (minimum), and radiation have the highest predictive importance shown by the normalized Shapley values of 13.20, 12.03, and 9.56, respectively. These observations make sense from an agronomic perspective and are in line with the findings of other studies suggesting temperature-related indicators having a high crop yield prediction power[53]. The higher Shapley values for minimum temperature and wind speed compared to the other weather components around weeks 35–37 coincide with the reproductive phase. This stage is the most sensitive growth stage to low temperatures and wind speed, affecting photosynthesis by changes in stomatal conductance and leaf temperature[54].

Based on Fig. 7, minimum and maximum temperature are among the top features contributing to yield prediction outputs. Several studies have reported that increased minimum temperature reduces the cold and frost induced damage to the seedlings[55,56], leading to increased crop yield and warming-induced increases in the crop grain weight[57]. Based on an agronomic point of view, the higher Shapley values of minimum temperature around week 31 compared to other weather components (Fig. S6) coincide with the flowering phase. On the other hand, a high temperature can negatively affect winter wheat grain numbers, grain size[58,59] and the duration of grain filling period [60], resulting in reduced crop yields. Furthermore, the adverse yield effects of high temperatures are also associated with the water and heat stress conditions around the sensitive stages of crop growth period. This may lead to reduced photosynthesis rate, increased respiration, higher rate of leaf senescence, and evapotranspiration, eventually reducing the grain numbers[61]. Precipitation holds an equally crucial predictive factor for winter wheat and the crop yield in general. This could be attributed to the fact that crops are exposed to drought stress under less precipitation, especially during the reproduction phase. Thus, grain yields can be negatively affected due to obstructed nutrient uptake, and in conjunction with reduced surface cooling, crop canopy temperatures shoot up leading to the further decline in photosynthetic rates[62]. Higher Shapley values of precipitation compared to the other weather variables around week 25, a period that could be linked to stem elongation, indicated higher predictive importance of this variable around this time. Stem elongation is a critical phase for yield formation and water is crucial for expansive growth processes of the crop, cell expansion, pollen ripening, spike growth and grain filling[63]. In terms of soil variables (refer to Fig. 7 and 8), our analysis showed DUL, SAT, and BD are important for yield prediction. This observation alig ns with Peichl et al.'s[64] findings, stating that soil moisture is the most critical variable for predicting winter wheat. The studied soil factors are known to affect the available crop water, which directly impacts growth and development. For insta nce, sandy soils with a lower water holding capacity may pose water stress conditions during the winter wheat growth period hampering the yield[65]. On the other hand, excessive water in the soil (waterlogged conditions) could lead to aeration stress, damaging the plants and increasing yield losses[46]. It was also reported that excessive soil water encourages the development of pathogens and hurdles the necessary management operations[67].

| | CNN (full model) | CNN (top 75%) | CNN (top 50%) | CNN (weather) |
|---|---|---|---|---|
| Test on 2017 (RMSE) | 0.66 | 0.71 | 0.70 | 1.20 |
| Test on 2018 (RMSE) | 0.84 | 0.75 | 0.96 | 1.11 |
| Test on 2019 (RMSE) | 0.64 | 0.80 | 0.81 | 1.15 |
| Test on 2017 (MAE) | 0.52 | 0.57 | 0.56 | 1.04 |
| Test on 2018 (MAE) | 0.71 | 0.61 | 0.77 | 0.90 |
| Test on 2019 (MAE) | 0.50 | 0.63 | 0.63 | 0.95 |
| Test on 2017 (r) | 0.65 | 0.51 | 0.55 | 0.35 |
| Test on 2018 (r) | 0.78 | 0.71 | 0.43 | 0.10 |
| Test on 2019 (r) | 0.81 | 0.67 | 0.63 | 0.50 |

**Table 2.** Test results of using different subsets of features for predicting winter wheat yield with the proposed CNN model.

*Feature selection.* For evaluating the performance of the feature importance metric (Shapley values), we studied the prediction results using an importance-based feature selection technique. As such, we ranked features based on their Shapley values and selected 75 and 50 percent of the most important features for the analysis. Table 2 shows the yield prediction performance of the proposed CNN model with the mentioned feature selection thresholds. Based on the results, the performance of the proposed CNN model not only did not drop significantly as a result of feature selection, but also got slightly improved, which confirms that the SHAP explanation method can effectively find the important features in the winter wheat yield prediction. In addition, the proposed CNN model was retrained using only the weekly weather variables to evaluate the yield prediction performance solely using weather features.

## Conclusions

In this paper, we proposed a convolutional neural network (CNN) model with 1-D convolution operations to predict the winter wheat yield using the weather, soil, and crop phenology data across 271 counties in Germany over 21 years (1999 to 2019). We used eight supervised machine learning models, including KNN, Random Forest, XGBoost, Regression Tree, Lasso and Ridge Regressions, SVR, and DNN, as baselines to compare the performance of our proposed model with them. Based on the chosen validation metrics (RMSE, MAE, and correlation coefficient), the results of our analysis showed that nonlinear models such as DNN, CNN, and XGBoost outperform linear models to a varying extent because crop yield is a highly complicated feature that depends on numerous interactive factors such as genotype and environmental conditions. Thus, nonlinear models are more effective in finding the functional relationship between the crop yield and input data. The results also revealed that our proposed CNN model outperforms all baselines across the test years of 2017, 2018, and 2019. The proposed CNN model has 7 to 14% lower RMSE, 3 to 15% lower MAE, and 4 to 50% higher correlation coefficient than the best performing baseline across the test years. We also moved beyond prediction in our analysis and measured feature importance using Shapley values on the test data to explain the important features in predictions made by our proposed black box CNN model. Furthermore, we visualized how changing the values of the underlying features will increase/decrease the winter wheat yield prediction values and identified the important ranges of values in each feature and their contribution to the predicted yield outputs. Also, we provided instance-level force plot visualizations and studied the important features, and their respective values for the instances at various predicted yield levels (the highest, lowest, and median). This information helps evaluate the important features at the middle and both ends of the prediction distribution. Additionally, we estimated the contribution of the weather components (precipitation, minimum and maximum temperature, radiation, relative humidity, and wind speed) across different weeks of the year to the winter wheat yield prediction results. The provided information is a helpful asset for farmers and agricultural scientists to understand the important time periods of weather components for yield values. Finally, we sorted the entire set of weather, soil, and phenological features based on their estimated effects on the prediction results (Shapley values) and conducted feature selection on them; we then retrained the proposed CNN model using the top 75 and 50% of features and also using weather features only and recorded the test results. This helped us to assess the performance of our feature importance evaluation metric (Shapley values); we found that the performance of the proposed CNN model did not drop drastically when using only a subset of the most important features compared to using the entire feature set, confirming the robustness of our feature importance method. The primary limitation of the proposed CNN model in this research is its black box nature; we could alleviate the lack of transparency in predictions made by our proposed black box model by using post-hoc explanation tools such as SHAP. However, some mathematical problems are identified about using Shapley values as feature importance tools (i.e., causal reasoning) [68]; also they are not considered a natural explanation tool based on human reasoning [68]. Explaining black box models using surrogate tools instead of building inherently-interpretable models might not be a sound scientific practice [69]. Thus, the future direction of this research is to build more advanced models that are both highly accurate and inherently interpretable.

## References

1. Konduri, V. S., Vandal, T. J., Ganguly, S. & Ganguly, A. R. Data science for weather impacts on crop yield. *Front. Sustain. Food Syst.* **4**, 52. https://doi.org/10.3389/fsufs.2020.00052 (2020).
2. Xu, X., Gao, P., Zhu, X., Guo, W., Ding, J., Li, C., ... & Wu, X. Design of an integrated climatic assessment indicator (ICAI) for wheat production: A case study in Jiangsu Province, China. *Eco. Ind.*, **101**, 943953. https://doi.org/10.1016/J.ECOLIND.2019.01.059 (2019).
3. Moeinizade, S., Hu, G., Wang, L. & Schnable, P. S. Optimizing selection and mating in genomic selection with a look-ahead approach: An operations research framework. *G3: Genes Genomes Genet.* **9**, 2123–2133. https://doi.org/10.1534/g3.118.200842 (2019).
4. Basso, B. & Liu, L. Chapter four: Seasonal crop yield forecast: Methods, applications, and accuracies. In *Advances in Agronomy* Vol. 154 (ed. Sparks, D. L.) 201–255. https://doi.org/10.1016/bs.agron.2018.11.002 (2019)
5. Zarindast, A., & Wood, J. A Data-Driven Personalized Lighting Recommender System. *Front. in Big Data*, **4**. (2021).
6. Shahhosseini, M., Hu, G., Khaki, S., & Archontoulis, S. V. Corn yield prediction with ensemble CNN-DNN. *Front. Plant Sci.*, **12**. (2021).
7. Haghighat, A. K., Ravichandra-Mouli, V., Chakraborty, P., Esfandiari, Y., Arabi, S., & Sharma, A. Applications of deep learning in intelligent transportation systems. *J. Big Data Anal. Transp.* **2**, 115–145. https://doi.org/10.1007/s42421-020-00020-1 (2020).
8. Zarindast, A., Sharma, A., & Wood, J. Application of text mining in smart lighting literature-an analysis of existing literature and a research agenda. *Int. J. Info. Mgmt. Data Insights*, **1**(2), 100032. (2021).
9. Chlingaryan, A., Sukkarieh, S. & Whelan, B. Machine learning approaches for crop yield prediction and nitrogen status estimation in precision agriculture: A review. *Comput. Electron. Agric.* **151**, 61–69. https://doi.org/10.1016/j.compag.2018.05.012 (2018).
10. Zarindast, A., Poddar, S., & Sharma, A. A Data-Driven Method for Congestion Identification and Classification. *J. Trans. Eng., Part A: Sys.*, **148**(4), 04022012. (2022)
11. Shahhosseini, M., Martinez-Feria, R. A., Hu, G. & Archontoulis, S. V. Maize yield and nitrate loss prediction with machine learning algorithms. *Environ. Res. Lett.* **14**, 124026. https://doi.org/10.1088/1748-9326/ab5268 (2019).
12. Khaki, S. & Wang, L. Crop yield prediction using deep neural networks. *Front. Plant Sci.*https://doi.org/10.3389/fpls.2019.00621 (2019).
13. Feng, P., Wang, B., Li Liu, D., Waters, C., Xiao, D., Shi, L., & Yu, Q. Dynamic wheat yield forecasts are improved by a hybrid approach using a biophysical model and machine learning technique. *Agric. For. Meteorol.* **285–286**, 107922. https://doi.org/10.1016/j.agrformet.2020.107922 (2020).
14. Kang, Y., Ozdogan, M., Zhu, X., Ye, Z., Hain, C., & Anderson, M. Comparative assessment of environmental variables and machine learning algorithms for maize yield prediction in the US Midwest. *Environ. Res. Lett.* **15**, 064005. https://doi.org/10.1088/1748-9326/ab7df9 (2020).
15. Van Klompenburg, T., Kassahun, A. & Catal, C. Crop yield prediction using machine learning: A systematic literature review. *Comput. Electron. Agric.* **177**, 105709. https://doi.org/10.1016/j.compag.2020.105709 (2020).
16. Khaki, S., Safaei, N., Pham, H. & Wang, L. *WheatNet: A Lightweight Convolutional Neural Network for High-throughput Image-based Wheat Head Detection and Counting. arXiv:2103.09408* (2021).
17. Hassan, M. A., Khalil, A., Kaseb, S. & Kassem, M. A. Exploring the potential of tree-based ensemble methods in solar radiation modeling. *Appl. Energy* **203**, 897–916. https://doi.org/10.1016/j.apenergy.2017.06.104 (2017).
18. Mishra, S. & Santra, D. M. A. G. H. Applications of machine learning techniques in agricultural crop production: A review paper. *Indian J. Sci. Technol.* **9**, 1–14. https://doi.org/10.17485/ijst/2016/v9i38/95032 (2016).
19. Kamilaris, A. & Prenafeta-Boldú, F. Deep learning in agriculture: A survey. *Comput. Electron. Agric.*https://doi.org/10.1016/j.compag.2018.02.016 (2018).
20. Liakos, K. G., Busato, P., Moshou, D., Pearson, S. & Bochtis, D. Machine learning in agriculture: A review. *Sensors* **18**, 2674. https://doi.org/10.3390/s18082674 (2018).
21. Wang, Y., Zhang, Z., Feng, L., Du, Q. & Runge, T. Combining multi-source data and machine learning approaches to predict winter wheat yield in the conterminous United States. *Remote Sens.* **12**, 1232. https://doi.org/10.3390/rs12081232 (2020).
22. Cao, J., Zhang, Z., Luo, Y., Zhang, L., Zhang, J., Li, Z., & Tao, F. Wheat yield predictions at a county and field scale with deep learning, machine learning, and google earth engine. *Eur. J. Agron.* **123**, 126204. https://doi.org/10.1016/j.eja.2020.126204 (2021).
23. FAO. FAOSTAT (2021).
24. Zhao, G., Webber, H., Hoffmann, H., Wolf, J., Siebert, S., & Ewert, F. The implication of irrigation in climate change impact assessment: a European-wide study. *Glob. Chang. Biol.* **21**, 4031–4048. https://doi.org/10.1111/gcb.13008 (2015).

25. EUROSTAT. Glossary: Nomenclature of territorial units for statistics (NUTS) (2019).
26. COPERNICUS. CORINE Land Cover (2006).
27. Webber, H., Lischeid, G., Sommer, M., Finger, R., Nendel, C., Gaiser, T., & Ewert, F. No perfect storm for crop yield failure in Germany. *Environ. Res. Lett.* **15**, 104012 (2020).
28. EUROSTAT. NUTS - Nomenclature of territorial units for statistics - Eurostat (2019).
29. Ämter, S. Regionaldatenbank Deutschland (2020).
30. DWD. Wetter und Klima - Deutscher Wetterdienst (2020).
31. Shook, J., Gangopadhyay, T., Wu, L., Ganapathysubramanian, B., Sarkar, S., & Singh, A. K. Crop yield prediction integrating geno-type and weather variables using deep learning. *PLOS ONE* **16**, e0252402. https://doi.org/10.1371/journal.pone.0252402 (2021).
32. Nevavuori, P., Narra, N., Linna, P. & Lipping, T. Crop yield prediction using multitemporal UAV data and spatio-temporal deep learning models. *Remote Sens.* **12**, 4000. https://doi.org/10.3390/rs12234000 (2020).
33. Breiman, L. Random forests. *Mach. Learn.* **45**, 5–32. https://doi.org/10.1023/A:1010933404324 (2001).
34. Cover, T. & Hart, P. Nearest neighbor pattern classification. *IEEE Trans. Inf. Theory* **13**, 21–27. https://doi.org/10.1109/TIT.1967.1053964 (1967).
35. Hu, L.-Y., Huang, M.-W., Ke, S.-W. & Tsai, C.-F. The distance function effect on k-nearest neighbor classification for medical datasets. *SpringerPlus* **5**, 1304. https://doi.org/10.1186/s40064-016-2941-7 (2016).
36. James, G., Witten, D., Hastie, T. & Tibshirani, R. *An Introduction to Statistical Learning: With Applications in R* (Springer, 2013).
37. Hoerl, A. E. & Kennard, R. W. Ridge regression: Biased estimation for nonorthogonal problems. *Technometrics* **12**, 55–67. https://doi.org/10.1080/00401706.1970.10488634 (1970).
38. Breiman, L., Friedman, J., Stone, C. J. & Olshen, R. *Classification and Regression Trees* 1st edn. (Chapman and Hall/CRC Press, 1984).
39. Cortes, C. & Vapnik, V. Support-vector networks. *Mach. Learn.* **20**, 273–297. https://doi.org/10.1007/BF00994018 (1995).
40. Chen, T., & Guestrin, C. Xgboost: A scalable tree boosting system. In *Proceedings of the 22nd acm sigkdd international conference on knowledge discovery and data mining* (pp. 785–794) (2016).
41. Zhang, L., Zhang, Z., Luo, Y., Cao, J. & Tao, F. Combining optical, fluorescence, thermal satellite, and environmental data to predict county-level maize yield in China using machine learning approaches. *Remote Sens.* **12**, 21. https://doi.org/10.3390/rs12010021 (2020).
42. Nigam, A., Garg, S., Agrawal, A. & Agrawal, P. Crop Yield Prediction Using Machine Learning Algorithms. *2019 Fifth International Conference on Image Information Processing (ICIIP)* https://doi.org/10.1109/ICIIP47207.2019.8985951 (2019).
43. LeCun, Y., Bengio, Y. & Hinton, G. Deep learning. *Nature* **521**, 436–444. https://doi.org/10.1038/nature14539 (2015).
44. Khaki, S., Wang, L. & Archontoulis, S. V. A CNN-RNN framework for crop yield prediction. *Front. Plant Sci.* https://doi.org/10.3389/fpls.2019.01750 (2020).
45. Khaki, S., Pham, H., & Wang, L. Simultaneous corn and soybean yield prediction from remote sensing data using deep transfer learning. *Sci. Rep.* **11**(1), 1–14. (2021).
46. Khaki, S., Khalilzadeh, Z. & Wang, L. Predicting yield performance of parents in plant breeding: A neural collaborative filtering approach. *PLoS ONE* **15**, e0233382. https://doi.org/10.1371/journal.pone.0233382 (2020).
47. Lamorski, K., Pachepsky, Y., Sławiński, C. & Walczak, R. T. Using support vector machines to develop pedotransfer functions for water retention of soils in Poland. *Soil Sci. Soc. Am. J.* **72**, 1243–1247. https://doi.org/10.2136/sssaj2007.0280N (2008).
48. Merdun, H., Çınar, C., Meral, R. & Apan, M. Comparison of artificial neural network and regression pedotransfer functions for prediction of soil water retention and saturated hydraulic conductivity. *Soil Tillage Res.* **1–2**, 108–116. https://doi.org/10.1016/j.still.2005.08.011 (2008).
49. Landeras, G., Ortiz-Barredo, A. & López, J. J. Comparison of artificial neural network models and empirical and semi-empirical equations for daily reference evapotranspiration estimation in the Basque Country (Northern Spain). *Agric. Water Manag.* **95**, 553–565. https://doi.org/10.1016/j.agwat.2007.12.011 (2008).
50. Yamaç, S. S. & Todorovic, M. Estimation of daily potato crop evapotranspiration using three different machine learning algorithms and four scenarios of available meteorological data. *Agric. Water Manag.* **228**, 105875. https://doi.org/10.1016/j.agwat.2019.105875 (2020).
51. Obsie, E. Y., Qu, H. & Drummond, F. Wild blueberry yield prediction using a combination of computer simulation and machine learning algorithms. *Comput. Electron. Agric.* **178**, 105778. https://doi.org/10.1016/j.compag.2020.105778 (2020).
52. Shapley, L. S. *A Value for n-Person Games* (Princeton University Press, 1953).
53. Vogel, E., Donat, M. G., Alexander, L. V., Meinshausen, M., Ray, D. K., Karoly, D., ... & Frieler, K. The effects of climate extremes on global agricultural yields. *Environ. Res. Lett.* **14**, 054010. https://doi.org/10.1088/1748-9326/ab154b (2019).
54. Stokes, V. J., Morecroft, M. D. & Morison, J. I. L. Boundary layer conductance for contrasting leaf shapes in a deciduous broadleaved forest canopy. *Agric. For. Meteorol.* **139**, 40–54. https://doi.org/10.1016/j.agrformet.2006.05.011 (2006).
55. Chen, X. & Chen, S. China feels the heat: Negative impacts of high temperatures on China's rice sector. *Aust. J. Agric. Resour. Econ.* **62**, 576–588. https://doi.org/10.1111/1467-8489.12267 (2018).
56. Tao, F., Xiao, D., Zhang, S., Zhang, Z. & Rötter, R. P. Wheat yield benefited from increases in minimum temperature in the Huang–Huai–Hai Plain of China in the past three decades. *Agric. For. Meteorol.* **239**, 1–14. https://doi.org/10.1016/j.agrformet.2017.02.033 (2017).
57. Zheng, C., Zhang, J., Chen, J., Chen, C., Tian, Y., Deng, A., ... & Zhang, W. Nighttime warming increases winter-sown wheat yield across major Chinese cropping regions. *Field Crops Res.* **214**, 202–210. https://doi.org/10.1016/j.fcr.2017.09.014 (2017).
58. Gibson, L. R. & Paulsen, G. M. Yield components of wheat grown under high temperature stress during reproductive growth. *Crop Sci.* **39**, 1841–1846. https://doi.org/10.2135/cropsci1999.3961841x (1999).
59. Lobell, D. B., Sibley, A. & Ivan Ortiz-Monasterio, J. Extreme heat effects on wheat senescence in India. *Nat. Clim. Chang.* **2**, 186–189. https://doi.org/10.1038/nclimate1356 (2012).
60. Tashiro, T. & Wardlaw, I. A comparison of the effect of high temperature on grain development in wheat and rice. *Ann. Bot.* **64**, 59–65. https://doi.org/10.1093/oxfordjournals.aob.a087808 (1989).
61. Wollenweber, B., Porter, J. R. & Schellberg, J. Lack of interaction between extreme high-temperature events at vegetative and reproductive growth stages in wheat. *J. Agron. Crop Sci.* **189**, 142–150. https://doi.org/10.1046/j.1439-037X.2003.00025.x (2003).
62. Mäkinen, H., Kaseva, J., Trnka, M., Balek, J., Kersebaum, K. C., Nendel, C., ... & Kahiluoto, H. Sensitivity of European wheat to extreme weather. *Field Crops Res.,* **222**, 209–217. https://doi.org/10.1016/j.fcr.2017.11.008 (2018).
63. Farooq, M., Bramley, H., Palta, J. A. & Siddique, K. H. Heat stress in wheat during reproductive and grain-filling phases. *Crit. Rev. Plant Sci.* **30**, 491–507. https://doi.org/10.1080/07352689.2011.615687 (2011).
64. Peichl, M., Thober, S., Meyer, V. & Samaniego, L. The effect of soil moisture anomalies on maize yield in Germany. *Nat. Hazards Earth Syst. Sci.* **18**, 889–906. https://doi.org/10.5194/nhess-18-889-2018 (2018).
65. Rezaei, E. E. *et al.* Quantifying the response of wheat yields to heat stress: The role of the experimental setup. *Field Crops Res.,* **217**, 93–103. https://doi.org/10.1016/j.fcr.2017.12.015 (2018).
66. Cannell, R. Q., Belford, R. K., Gales, K., Dennis, C. W. & Prew, R. D. Effects of waterlogging at different stages of development on the growth and yield of winter wheat. *J. Sci. Food Agric.* **31**, 117–132. https://doi.org/10.1002/jsfa.2740310203 (1980).
67. Gömann, H. Wetterextreme: mögliche Folgen für die Landwirtschaft in Deutschland (2018).

68. Kumar, I. E., Venkatasubramanian, S., Scheidegger, C., & Friedler, S. Problems with Shapley-value-based explanations as feature importance measures. In International Conference on Machine Learning (pp. 5491–5500). *PMLR*. (2020).
69. Rudin, C. Stop explaining black box machine learning models for high stakes decisions and use interpretable models instead. *Nature Machine Intelligence*, **1**(5), 206–215. (2019).

## Acknowledgements

The presented study has been funded by the German Federal Ministry of Education and Research (BMBF) in the framework of the funding measure 'Soil as a Sustainable Resource for the Bioeconomy— BonaRes', project BonaRes (Module A): BonaRes Center for Soil Research, subproject 'Sustainable Subsoil Management—Soil3' (Grant 031B0151A), and partially funded by the Deutsche Forschungsgemeinschaft (DFG, German Research Foundation) under Germany's Excellence Strategy—EXC 2070—390732324.

## Author contributions

A.S. and S.K. conceptualised the research. N.S. and S.K. contributed in model development and data analysis. A.S., S.K., and N.S. wrote the manuscript. All authors contributed to the article and approved the submitted version.

## Funding

Open Access funding enabled and organized by Projekt DEAL.

## Competing interests

The authors declare no competing interests.

## Additional information

**Supplementary Information** The online version contains supplementary material available at https://doi.org/10.1038/s41598-022-06249-w.

**Correspondence** and requests for materials should be addressed to A.K.S., N.S. or S.K.

**Reprints and permissions information** is available at www.nature.com/reprints.

**Publisher's note** Springer Nature remains neutral with regard to jurisdictional claims in published maps and institutional affiliations.